\documentclass[runningheads]{llncs}
\usepackage{graphicx}
\usepackage{tikz}
\usepackage{comment}
\usepackage{amsmath,amssymb}
\usepackage{color}
\usepackage[accsupp]{axessibility} 

\usepackage{algorithm}
\usepackage{listings}
\usepackage{booktabs}
\usepackage{multirow}
\usepackage{pifont}

\usepackage{float}
\usepackage{subfig}
\usepackage{wrapfig}

\usepackage{marvosym}

\usepackage[pagebackref,breaklinks,colorlinks]{hyperref}

\newcommand{\cmark}{\ding{51}}
\newcommand{\xmark}{\ding{55}}
\newcommand{\etal}{\textit{et al}.}
\newcommand{\ie}{\textit{i}.\textit{e}. }
\newcommand{\eg}{\textit{e}.\textit{g}. }
\newcommand{\wrt}{\textit{w}.\textit{r}.\textit{t}. }
\newcommand{\etc}{\textit{etc}}

\usepackage[capitalize]{cleveref}
\crefname{section}{Sec.}{Secs.}
\Crefname{section}{Section}{Sections}
\Crefname{table}{Table}{Tables}
\crefname{table}{Tab.}{Tabs.}

\newcommand\figcaption{\def\@captype{figure}\caption}
\newcommand\tabcaption{\def\@captype{table}\caption}

\usepackage[width=122mm,left=12mm,paperwidth=146mm,height=193mm,top=12mm,paperheight=217mm]{geometry}

\newcommand\blfootnote[1]{
  \begingroup
  \renewcommand\thefootnote{}\footnote{#1}
  \addtocounter{footnote}{-1}
  \endgroup
}

\begin{document}
\pagestyle{headings}
\mainmatter

\title{Logit Normalization for Long-tail Object Detection}

\titlerunning{Logit Normalization for Long-tail Object Detection}
\author{Liang Zhao\textsuperscript{*} \and
Yao Teng\textsuperscript{*} \and
Limin Wang\textsuperscript{\Letter}
}
\authorrunning{L. Zhao, Y. Teng, L. Wang}
\institute{State Key Laboratory for Novel Software Technology, Nanjing University, China \\
\email{liangzhao@smail.nju.edu.cn, tengyao19980325@gmail.com, lmwang@nju.edu.cn}}
\maketitle

\begin{abstract}
Real-world data exhibiting skewed distributions pose a serious challenge to existing object detectors. Moreover, the samplers in detectors lead to shifted training label distributions, while the tremendous proportion of background to foreground samples severely harms foreground classification. To mitigate these issues, in this paper, we propose Logit Normalization (LogN), a simple technique to self-calibrate the classified logits of detectors in a similar way to batch normalization. In general, our LogN is training- and tuning-free (\ie require no extra training and tuning process), model- and label distribution-agnostic (\ie generalization to different kinds of detectors and datasets), and also plug-and-play (\ie direct application without any bells and whistles). Extensive experiments on the LVIS dataset demonstrate superior performance of LogN to state-of-the-art methods with various detectors and backbones. We also provide in-depth studies on different aspects of our LogN. Further experiments on ImageNet-LT reveal its competitiveness and generalizability. Our LogN can serve as a strong baseline for long-tail object detection and is expected to inspire future research in this field. Code and trained models will be publicly available at \url{https://github.com/MCG-NJU/LogN}.
\end{abstract}
\blfootnote{\small *: Equal contribution. \Letter: Corresponding author.}

\vspace{-5mm}
\section{Introduction}
\label{sec:intro}

Large-scale datasets collected manually such as MS COCO~\cite{coco} and Open Images~\cite{openimage} have delivered tremendous success on object detection.
However, in comparison with data from realistic scenarios, they tend to be relatively balanced, 
since realistic data often exhibit skewed distributions with a long tail~\cite{longtail_intro1,imagenetlt}.
In this sense, a handful of head (frequent) classes containing massive samples dominate,
yet a lot of tail (rare) classes containing a few samples are thus concealed.
Such a class imbalance poses considerable challenges for object detectors.

\begin{figure}[!tb]
    \centering
    \subfloat[Differences between logarithms of training label distributions (with and without external sampler,  RFS~\cite{lvis}) and logarithm of prior label distribution.]
    {
        \includegraphics[width=0.44\textwidth]{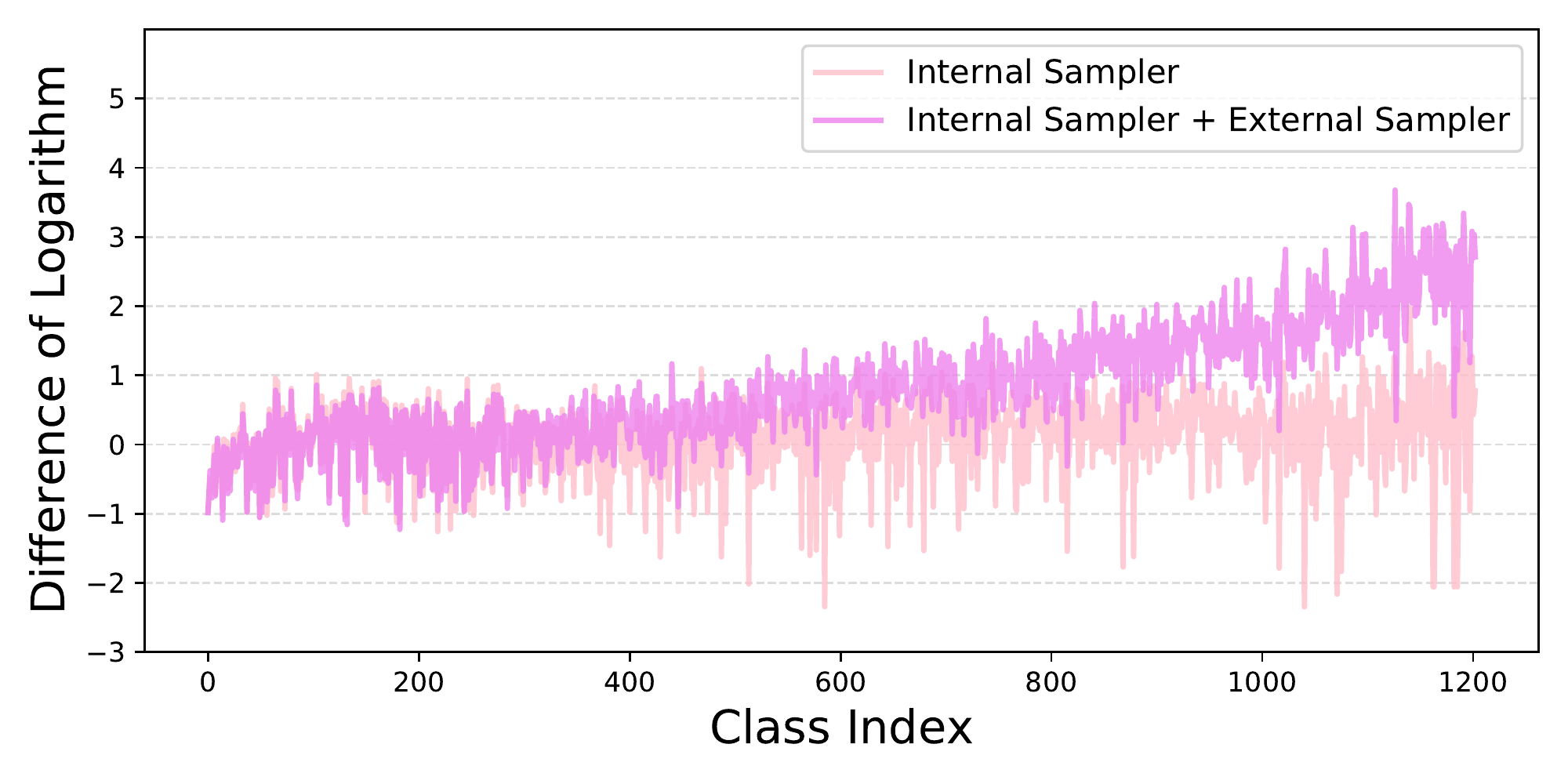}
        \label{subfig1:prior_label}
    }
    \qquad
    \subfloat[Per and cumulative ratios of the number of samples in foreground class ($N_{FG}$) relative to the number of samples in the background class ($N_{BG}$).]
    {
        \includegraphics[width=0.44\textwidth]{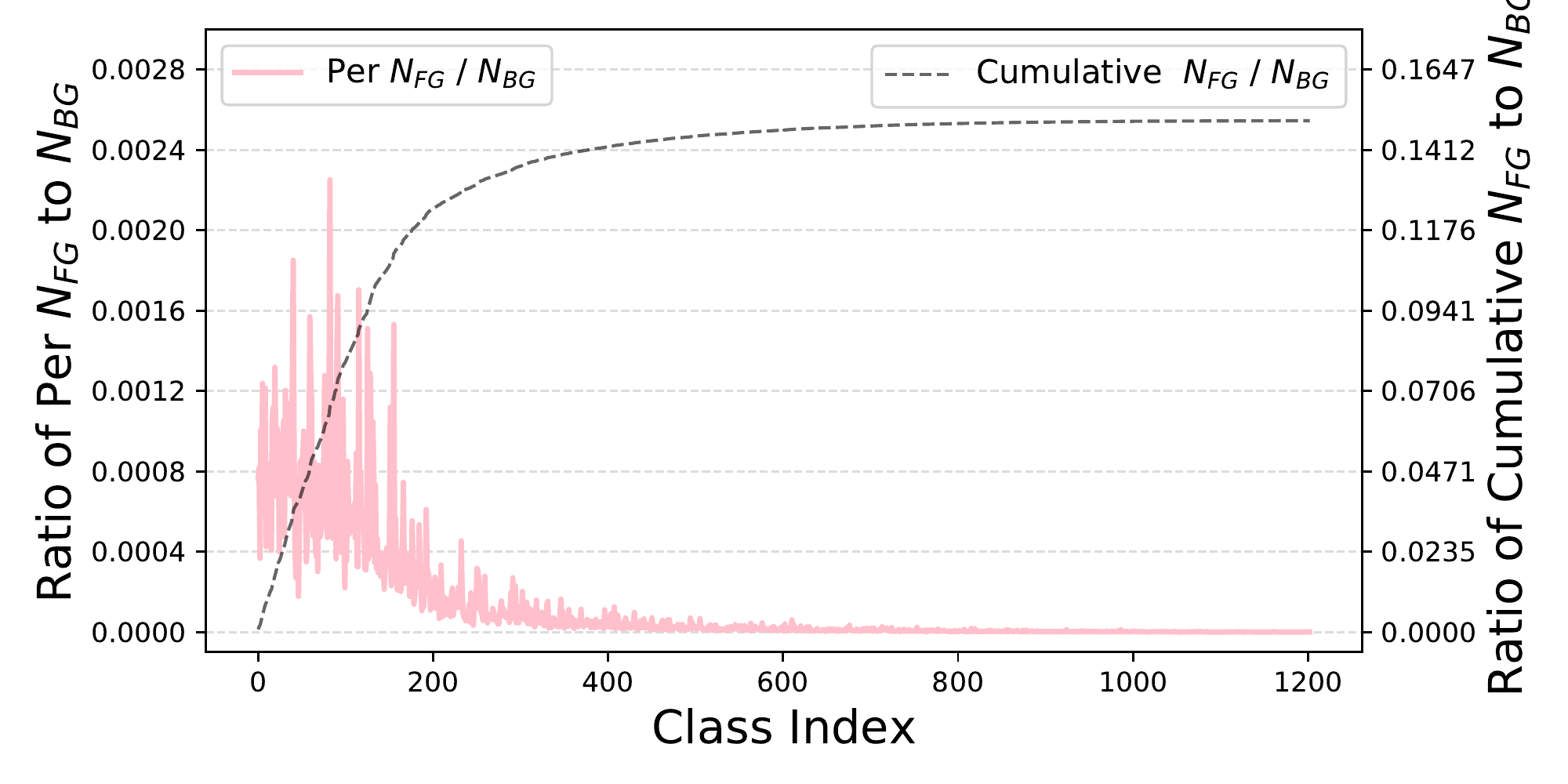}
        \label{subfig1:background}
    }
    \vspace{-1mm}
    \caption{
    The x-axis indicates the category index sorted according to the number of samples included.
    (a) Relative to prior label distribution, internal sampler causes fluctuations and external sampler causes shifts, co-leading to the alternations of training label distributions.
    (b) Tremendous ratios of foreground to background samples from head to tail classes could influence the foreground classification.
    }
    \vspace{-3mm}
    \label{fig1}
\end{figure}

Apart from the skewed class distribution, we also summarize the other two challenges from the perspective of detection:
(1) The samplers in object detectors lead to a \textit{shift} between label frequencies~\cite{labelfreq,logit_adjustment} (\ie 
the frequency of samples in each category)
calculated in advance and those calculated during training, as shown in~\cref{subfig1:prior_label}.
In this paper, we term the frequency calculated in advance as prior label distribution and the frequency calculated during training as training label distribution.
Obviously, for one detector, its intrinsic label assignment~\cite{labelassign1,labelassign2}
which is based on an internal sampler, 
\textit{fluctuates} the original prior label distribution.
Furthermore, with the commonly attached external sampler such as RFS~\cite{lvis}, the final training label distribution is gradually \textit{shifted} from head to tail categories.
(2) The tremendous proportion of background proposals depicted in~\cref{subfig1:background} severely harms the capability of object detectors in classifying foreground proposals.
Unlike regular image classifiers, the classifier in object detector is required to precisely discriminate the foreground proposals from the background. 
However, for \textit{each} foreground category, the extreme foreground-background class imbalance~\cite{focal} severely affects its classification.

Based on these challenges, there are diverse techniques for tackling long-tail object detection~\cite{eqlv1,eqlv2,seesawloss,equilibrium_loss,norcal}.
Re-sampling~\cite{lvis}, 
an intuitive method that re-balances the number of training samples from head to tail classes,
has not gained enough favor in community. 
This technique not only leads to higher training complexity, but also easily causes the issues of under-fitting to head classes and over-fitting to tail classes.
Far more methods~\cite{eqlv1,balanced_meta_softmax,norcal} aim at re-weighting training loss by modifying the \textit{logits}, 
\ie the classified outputs, with class \textit{priors}, \eg the label frequency of training set.
Although these methods are theoretically proven to be well suited for classification tasks~\cite{balanced_meta_softmax}, unfortunately, they are often not well adapted to object detectors due to the shift caused by samplers~\cite{balanced_group_softmax}.
There are also several approaches~\cite{eqlv2,seesawloss,equilibrium_loss} that seek to dynamically accommodate long-tail distribution with \textit{statistics}, but they contain lots of parameters.
Moreover, in long-tail object detection, most methods lack an adaptive and unified design for the background class like the foreground classes.

\begin{figure}[!tb]
    \centering
    \subfloat[Logit Mean.]
    {
        \includegraphics[width=0.45\textwidth]{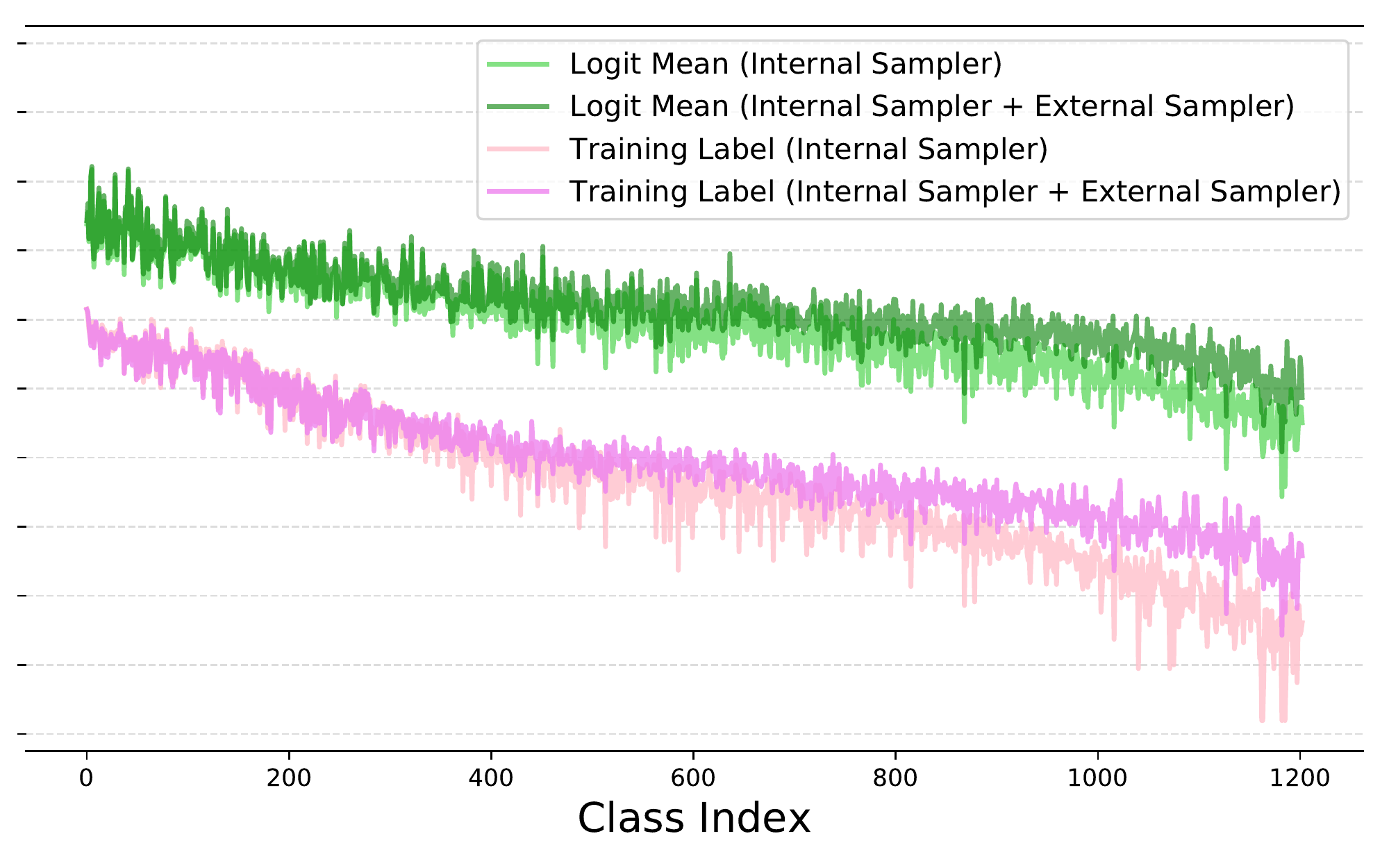}
        \label{subfig2:only_mu}
    }
    \subfloat[Logit Variance.]
    {
        \includegraphics[width=0.45\textwidth]{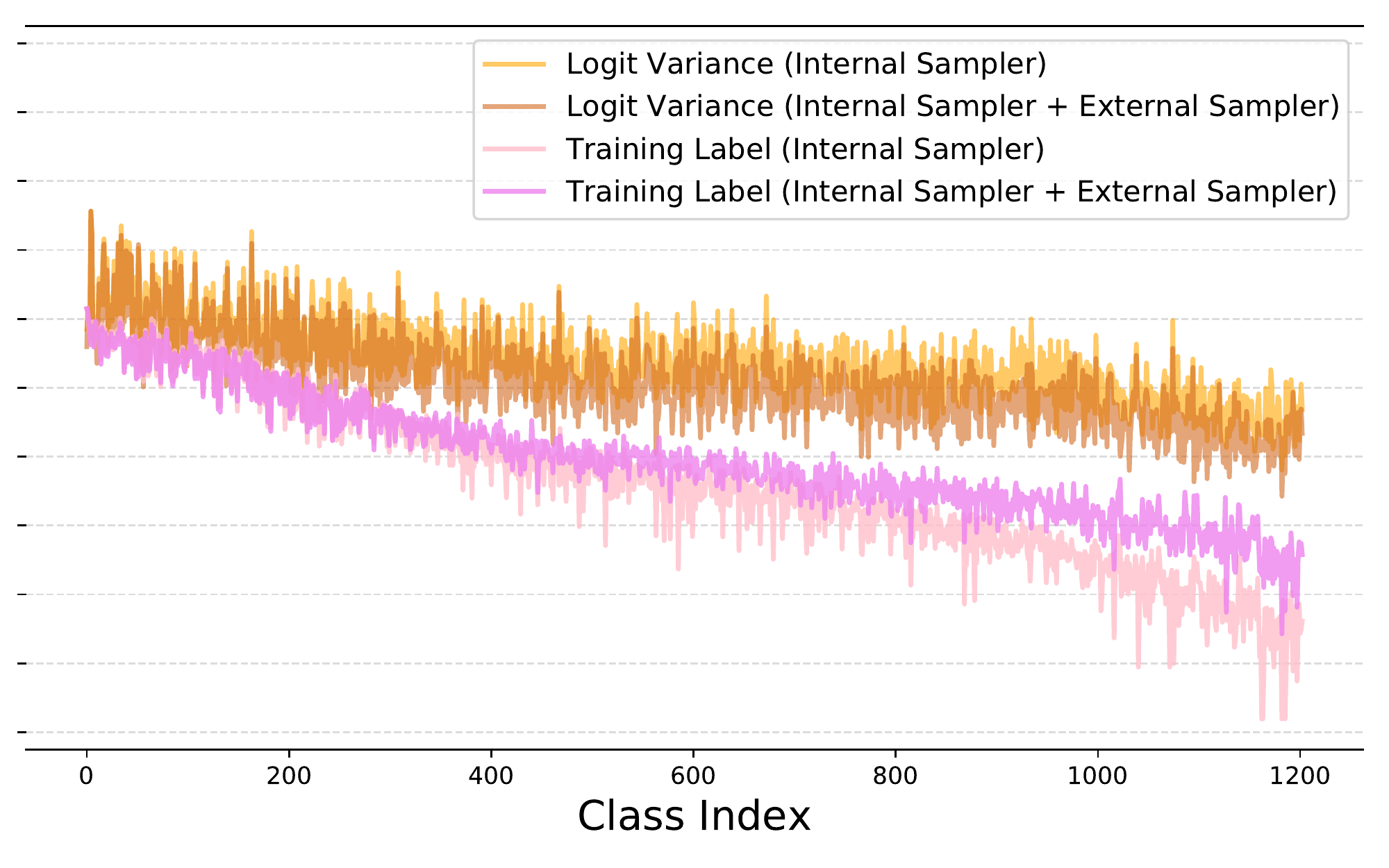}
        \label{subfig2:only_var}
    }
    \vspace{-2.5mm}
    \caption{Logit statistics and logarithms of training label distributions, with and without external sampler. (a) Logit mean consistently reflects the inter-class long-tail tendency of the dynamic training label distributions.
    (b) Logit variance is also correlated with the long-tail distribution.
    }
    \vspace{-.5cm}
    \label{mu_var}
\end{figure}

In this paper, we propose \textbf{Logit Normalization} (LogN), a conceptually simple yet quite effective \textit{post-hoc} calibration method for long-tail object detection.
By shifting the operation of batch normalization (BN) to the logits,
we are able to rectify the predictions of detectors biased by long-tail distribution, in a highly simple and versatile style.
Our LogN is based on two observations:
(1) The statistics (\ie mean and variance) at logits are capable of characterizing the inter-class long-tail tendency with label distribution, as illustrated in~\cref{mu_var}.
This phenomenon, in conjunction with the prior normalization-based work~\cite{in,cin,adain} for style translation,
motivates us to leverage the predictions themselves for self-calibration, so as to realize a sort of \textit{Long-tailedness Normalization}.
As a BN-like calibration method, it can be flexibly fitted to arbitrary detectors or distributions, without injecting any label-related priors.
(2) Improving the discriminability of classifier on background instead contributes to its foreground classification, as analyzed in~\cref{sec:analysis}.
On this basis, we increase the \textit{margin} between foreground and background logits via adding a negative scalar to the logit of each foreground class, so as to accomplish an equivalent \textit{background calibration}.
Practically, we provide an adaptive solution for this margin, which allows our LogN to be parameter-free.

Overall, LogN comprises the following three compelling characteristics: 
\textbf{(1)} \textbf{Training-free} and \textbf{tuning-free}:
as a post-hoc normalization technique, 
LogN requires no independent training as in~\cite{eqlv1,eqlv2,equilibrium_loss}, and also requires no extensive parametric adjustments as in~\cite{eqlv2,disalign}, being highly efficient and resource-saving.
\textbf{(2)} \textbf{Model-agnostic} and \textbf{label distribution-agnostic}:
when both the network structure and label distribution are invisible, LogN is still capable of delivering promising results, guaranteeing maximum security and privacy~\cite{privacy1} of the model and data compared to previous methods.
\textbf{(3)} \textbf{Plug-and-play}: 
it can be easily implemented with a few codes appended to the predicted logits,
and simply deployed to various platforms, without any bells and whistles. 

Extensive experiments have been conducted on the challenging long-tail object detection dataset, LVIS~\cite{lvis},
and the results demonstrate superior performance to the state-of-the-art models with various detectors including Mask R-CNN~\cite{maskrcnn} and Cascade Mask R-CNN~\cite{cascade},
using diverse backbones including ResNet-50-FPN~\cite{resnet,fpn} and ResNet-101-FPN.
We further experiment on the long-tail image classification dataset ImageNet-LT~\cite{imagenetlt},
and the competitive performance illustrates the favorable generality of LogN.

\section{Related Work}

\textbf{Normalization.}
Generally, normalization techniques can be divided into two groups, 
targeting \textit{weights}~\cite{weight1,weight2,weight3,weight4} and \textit{features}~\cite{bn,ghost,cbn,ln,gn,in,cin}.
This work is a generalization of feature normalization, thus we focus on the latter.

Ioffe and Szegedy~\cite{bn} pioneered a feature-level batch normalization (BN). 
By normalizing activations using feature statistics within a mini-batch, BN conditions the training with high learning rates and hence improves generalization accuracy effectively.
Afterwards, various subsequent methods shift their attention from the \textit{batch} dimension to 
(1) \textit{fine-grained batch} dimension: Cross-GPU BN~\cite{syncbn} on larger batch size across multiple GPUs, Ghost Normalization~\cite{ghost} on small virtual batch
and Cross-Iteration BN~\cite{cbn} on recent iterations of batch;
(2) \textit{channel} dimension: Layer Normalization~\cite{ln} on all channels, Group Normalization~\cite{gn} on groups of channels, and Instance Normalization (IN)~\cite{in} on each channel;
(3) \textit{combined} dimension: Switchable Normalization series~\cite{switch1,switch2,switch3} on mixtures of multiple dimensions including batch, channel, \etc.
In contrast to these methods that overly focus on intermediate features, 
our method operates normalization during inference on the logits for the first time, 
and is devoted to standardizing the biased predictions of detectors suffered from long-tail learning.

It also worth noting that IN and its subsequent methods (\eg Conditional IN~\cite{cin} and Adaptive IN~\cite{adain}) not only achieve impressive results for practical cross-domain learning, but also prompt a rethinking of normalization in terms of transfer learning.
Huang~\etal~\cite{adain} found that instance-wise normalization is essentially a kind of style purification, while the subsequent operation of scale and shift is the adaptation.
In this paper, we continue this thought, but focus on label distribution rather than on feature distribution that characterize styles. 
We purify biased predictions by simply normalizing them with logit statistics, enabling the detectors to better cope with relatively balanced test distribution.

\textbf{Long-Tail Learning.}
A great number of methods have been proposed for long-tail learning.
Lin~\etal~\cite{focal} proposed focal loss to mine the hardness of classification correlated with the label imbalance, and then online modified the penalty on each prediction.
Kang~\etal~\cite{decouple} utilized decoupled training strategy to disentangle the representation learning with the classifier training, and separately trained a balanced classifier to rectify the decision boundaries.
Logit Adjustment (LogA)~\cite{balanced_meta_softmax,logit_adjustment}
leveraged the label frequency to adjust logits and theoretically guaranteed its optimality \wrt the ideal image classification tasks.

Object detection suffers more severely from the long-tail category distribution~\cite{lvis}. Up to now, considerable methods have been proposed to address this challenge.
Repeat factor sampling~(RFS)~\cite{lvis} is a re-sampling strategy applied to this task in early research and its core is oversampling the rare categories. 
Balanced Group Softmax~(BAGS)~\cite{balanced_group_softmax} split the object categories into multiple groups and modified the application of softmax. 
Tan~\etal~\cite{eqlv1,eqlv2} proposed a series of equalization loss for balancing the weights of frequent and rare classes from a gradient competition view.
Pan~\etal~\cite{norcal} introduced directly 
LogA into long-tail object detection and kept the background score intact.
Recently, several efforts~\cite{eqlv2,equilibrium_loss,seesawloss} start to inject statistics into classification loss to dynamically accommodate the training distribution, for harmonizing the training of each class more accurately.
Among them, Seesaw loss~\cite{seesawloss} balanced the training on different categories via online accumulated instance numbers and predicted probability.
EQL v2~\cite{eqlv2} dynamically equalized gradients between positives and negatives with the gradient statistics.
Feng~\etal~\cite{equilibrium_loss} brought statistics of classification scores into long-tail foreground object classification and set a hyper-parameter to adjust the background score.
Problematically however, these methods typically exhibit complicated forms with lots of parameters. Moreover,
they overly focus on the imbalance within foreground object categories yet overlooked the appropriate treatment of background classification.
This paper utilizes a format of normalization to calibrate the logits in the phase of testing, ensuring both the efficiency and simplicity. We further analyze in detail the critical role of background on long-tail detectors and provide a concise and adaptive solution.

\section{Logit Normalization}

Most real-world data is long-tail in nature, and generally the number of samples per class diminishes exponentially from head to tail classes.
In this context, a general classifier based on random sampling suffers from a training label distribution that is highly \textit{consistent} with the long-tail prior label distribution, resulting in samples from head classes predominant and those from tail classes under-learned.
Thus a model, especially a detector trained on a long-tail training set, often underperforms on a class-balanced test set.
Various methods~\cite{balanced_meta_softmax,logit_adjustment,norcal} calibrate such a biased distribution by injecting explicitly the prior label distribution, or namely label frequencies into logits.
Despite great success in image classification, their performance in object detection is poorly guaranteed~\cite{balanced_group_softmax}.

Different from image classification, the training label distribution in object detection is normally \textit{inconsistent} with prior distribution, as illustrated in~\cref{subfig1:prior_label}.
It can be observed that there is always a difference between the prior and training label distribution.
We ascribe this phenomenon to the samplers within object detectors.
Regular detectors normally utilize the label assignment mechanism~\cite{labelassign1,labelassign2}, an implicit internal sampler,
for handling the label co-occurrence issue~\cite{survey}.
Yet this sampler implicitly oscillates label frequencies and hence renders a fluctuating training label distribution.
Additionally, some detectors further introduce external samplers~\cite{lvis,equilibrium_loss} that re-sample images or instances to alleviate severe long-tail bias, explicitly shifting the label distribution.
These samplers co-lead to an altered training label distribution.
Currently, in response to the shift of distribution, several recent methods~\cite{eqlv2,seesawloss,equilibrium_loss} propose to dynamically adapt, in terms of the \textit{statistics} on logit. 
They mostly specialize in the relevance between the label distribution and the mean of classification scores or logits, which consistently reflects the inter-class long-tail tendency, as shown in~\cref{subfig2:only_mu},
but ignore the more in-depth 2-order statistic, variance, which is also correlated with the training label distribution, as shown in~\cref{subfig2:only_var}.
Furthermore, logit variance is adequate to represent the intra-class magnitude~\cite{rmsprop,magnitude}, or namely the \textit{hardness}~\cite{variance_discrimination} of categories, and is able to balance the decision boundaries of classifiers like $\tau$-norm~\cite{decouple}.

These statistics (\ie the mean and variance) on features are often investigated in~\cite{bn,ln,gn} for normalizing activations.
They are also commonly considered as a surrogate to image style~\cite{in,cin,adain} and facilitate the practice of style normalization.
However, the association between them and the categories, or the label distribution
is still poorly explored.
Recently, Frosst \etal~\cite{snnl} found that networks typically persist in developing class-independent representations until the final layer, 
where class-dependent logits are exploited for discriminating, 
indicating a \textit{disagreement} between the statistics on features and logits \wrt the categories.
These works motivate us to explicitly develop the correlation between logit statistics and the label distribution.

Another critical issue in detection is the background class. 
Unlike image classifiers, a well-trained object detector is capable of distinguishing the foreground objects from the background. 
Generally, detectors treat the background as a specific \textit{background} category, 
and the built-in classifiers are usually designed with $C+1$ categories, where $C$ denotes the number of foreground object categories.
In long-tail object detection, the ability of object detectors in classifying foreground objects is always constrained by the proportion of background proposals.
As depicted in~\cref{subfig1:background}, the number of training instances of each foreground category is greatly less than those of the background.
For this problem, past practices either directly overlook the background class~\cite{eqlv1,eqlv2,norcal} or manually calibrate it with extra parameters~\cite{equilibrium_loss}.
A proper analytical and adaptive treatment towards the background class is urgently needed.

Inspired by the above,
we propose a novel post-hoc logit normalization for long-tail object detection.

\subsection{Formulation}

We first describe a general formulation of our logit normalization on the classification task, and then present its formulation when applied to object detection.
For the $C$-d logits output by a classifier, the normalization performs on the whole vector in the testing phase. Its general formulation is as follows:
\begin{equation}
    \hat{x} = \frac{x-\mu}{\sqrt{\sigma^2+\varepsilon }},
\label{Equ:bn1}
\end{equation}
where $x \in \mathbb{R}^C$ indicates the logits and $\mu,\sigma \in \mathbb{R}^C$ denote their mean and standard deviation post-hoc calculated from the training samples. $\varepsilon$ is equipped for numerical stability.
To efficiently estimate these statistics, we employ \textit{exponential moving average} in practice to accumulate global statistics from batch-wise statistics as in BN~\cite{bn}.
Notably, each statistic is calculated over \textbf{all} samples.

When it comes to the object detection task, networks first extract features for each ROI and then categorize them into batch of logits, 
$X \in \mathbb{R}^{N \times (C+1)}$, 
each of which comprises logits on $C$ foreground classes as well as one background class.
Contrary to the calibration in image classification, it is detrimental to object detection performance if we directly mimic this form of calibration on all classes.
In experiments, we find the reason behind this phenomenon is the statistics of the background category.
Thus, we additionally consider a \textit{background calibration} to repair the calibration on background.
Concretely, we use a specific scalar to consistently modify the logits of \textit{all} foreground classes, according to the equivalence of the operation on background logit and all foreground logits.
This operation can be viewed as imposing a certain \textit{margin} between each foreground logit and the background one. 
Therefore, the formulation of the normalization on long-tail object detection becomes as follows:
\begin{equation}
    \hat{x} = \frac{x-\mu+\beta}{\sqrt{\sigma^2+\varepsilon }},
\label{Equ:logn}
\end{equation}
where $\beta \in \mathbb{R} $ denotes the background calibration scalar,
and it can be assigned with the minimum value of the mean of foreground logits. $x \in \mathbb{R}^C$ indicates \textit{foreground} logits and $\mu,\sigma \in \mathbb{R}^C$ denote their mean and standard deviation.

In~\cref{alg:code}, we illustrate the PyTorch-like implementation on object detection of our LogN.
\textit{First}, we \textbf{offline} traverse the training set and aggregate the global statistics.
\textit{Second}, we equivalently perform the background calibration on all foreground logits.
\textit{Third}, we apply our LogN on classified logits ${x}$, and these transformed logits are for scores.
Since the overall algorithm is exclusively executed after training, we also refer to it as the \textbf{post-hoc} Logit Normalization.

\begin{algorithm}[t]
\caption{LogN Pseudocode, PyTorch-like}
\label{alg:code}
\definecolor{codeblue}{rgb}{0.25,0.5,0.5}
\definecolor{codekw}{rgb}{0.85, 0.18, 0.50}
\lstset{
  backgroundcolor=\color{white},
  basicstyle=\fontsize{7.5pt}{7.5pt}\ttfamily\selectfont,
  columns=fullflexible,
  breaklines=true,
  captionpos=b,
  commentstyle=\fontsize{7.5pt}{7.5pt}\color{codeblue},
  keywordstyle=\fontsize{7.5pt}{7.5pt}\color{codekw},
}
\begin{lstlisting}[language=python]
# detector: a well-trained object detector
# running_mean: global mean, running_var: global variance
# beta: background calibration scalar
# eps: 1e-5 for numerical stability, EMA: Exponential Moving Average

detector.eval() # the following steps are during **testing**

# 1. traverse training samples for calculating statistics
for x in train_loader: # load a training batch x
    feat = detector.extractor(x) # extract (ROI) features
    logit = detector.cls_head(feat) # output classifications
    mean = logit.mean(dim=0), var = logit.var(dim=0) # compute statistics in a mini-batch 
    
    # update global mean and variance with EMA 
    running_mean = EMA(running_mean, mean), running_var = EMA(running_var, var)

# 2. background calibration
beta = running_mean.min() # an adaptive beta value
running_mean = running_mean - beta
running_mean[0] = 0, running_var[0] = 1 # index 0 as the background

# 3. normalize logit for testing samples
for x in test_loader: # load a testing batch x
    feat = detector.extractor(x) # extract (ROI) features
    logit = detector.cls_head(feat) # output classifications
    normed_logit = (logit - running_mean) / sqrt(running_var + eps) # Logit Normalization
\end{lstlisting}
\vspace{-2mm}
\end{algorithm}

\subsection{Discussion}\label{s3}
 
\textbf{Relation to BN.}
The proposed LogN is formally close to BN~\cite{bn}, \ie they both use statistics for normalization.
This ensures the versatility of LogN,
since by post-hoc enforcing BN on logit, it can be conveniently adapted to any detectors and distributions in a plug-and-play manner,
without training or tuning.
However, differently,
BN normalizes intermediate activations with feature statistics to accelerate network training;
while LogN normalizes final activations with logit statistics to cope with long-tail learning.
Meanwhile, BN functions both in training and testing phases, to ensure the homogeneity in the input distribution of the data in both phases;
yet LogN functions solely in testing phase, to handle their heterogeneity in the label distribution.
Additionally, LogN does not include scale and shift, as the normalized data exhibit a basically uniform label distribution, which is also guaranteed in test set by a regular long-tail setting.

\textbf{Relation to Prior-based Methods.}
Existing methods~\cite{eqlv1,balanced_meta_softmax,logit_adjustment,norcal} focus on the application of dataset priors.
They typically inject label frequencies into the classified logits, hence fail in accommodating the re-sampled training label distribution within detectors.
In contrast, our LogN that normalizes logits with post-hoc calculated statistics is capable of dynamically adapting this altered distribution.
Moreover, the BN-like formulation renders LogN as a non-parametric self-calibration
approach, 
and thus is more flexible than the parametric prior-based methods such as LogA~\cite{logit_adjustment}.
Additionally, LogN also imposes calibration of background class, which is overlooked by previous prior-based methods.

\textbf{Relation to Statistic-based Methods.}
Several recent methods~\cite{eqlv2,seesawloss,equilibrium_loss} propose to accommodate the dynamic distribution within detectors
in terms of statistics. Despite the favorable performance achieved, most of them are formally complicated and overly parametric.
By contrast, our LogN enforces post-hoc BN on classified logits, calibrating the biased detections without any training or tuning procedure.
Moreover, existing practices~\cite{equilibrium_loss} rely heavily on a manual tuning hyper-parameter to adjust the weight of background logit.
In comparison, we analyze the processing of background logit and provide an adaptive choice for $\beta$ selection, 
sidestepping the tuning on the background logit calibration.
The experiments demonstrate the effectiveness of resilient background calibration.

\subsection{Analysis}\label{sec:analysis}

\textbf{Why Post-hoc?}\label{post-hoc}
Compared with the primitive BN, we remove the batch-wise normalization during training.
We conjecture that this is because
the normalization of the training phase increases the shift between the learned posterior distribution
and the training label distribution.
Specifically, the minimization of the objective function describing the consistency between these two distributions drives the former to adapt to the latter. 
Therefore, forcing the normalized posterior distribution to fit the long-tail bias will eventually lead to more biased predictions.
Two natural choices exist at this point, 
we can either delay the normalization until the testing phase (\ie our post-hoc LogN), or \textit{inverse} the normalization process in the training phase as follows:
\begin{equation}
    \hat{x} = x \cdot\sqrt{\sigma^2+\varepsilon } + \mu - \beta.
\label{Equ:invnorm}
\end{equation}
Here, the $\mu,\sigma$ are the persistent statistics accumulated over the training set, 
rather than the temporary batch-wise statistics used in BN.
This online alternative renders the modified posterior distribution more skewed and thereby easily adapted to the long-tail target distribution,
hence keeping the original posterior distribution more uniform. 
As indicated in~\cref{exp:analysis}, this online inverse LogN also provides relatively balanced predictions.

\textbf{Why with $\beta$?} 
Considering the case of no foreground calibration, in a $(C+1)$-d score vector, the score on the $i$-th category with background calibration is:
\begin{equation}
\begin{gathered}
\hat{s}_i = \frac{e^{x_i}}{e^{x_{0}-\beta} + \sum_{j=1}^{C}{e^{x_j}}} = \frac{e^{x_i+\beta}}{e^{x_{0}} + \sum_{j=1}^{C}{e^{x_j+\beta}}},
\end{gathered}
\label{Equ:bg_cal}
\end{equation}
where the 0th class is set as the background.
\cref{Equ:bg_cal} reveals that both the foreground and background logits are involved in the calculation of scores and they jointly affect the confidence ranking for the final evaluation~\cite{rank1,rank2}. Meanwhile, a single scalar $\beta$ is sufficient to calibrate the impact of the background, and 
this \textit{\textbf{background calibration} is equivalent to the consistent calibration on all foreground categories}.
We further reformulate the foreground score with logarithmic operation since it does not change the monotonicity:
\begin{equation}\small
\log \hat{s}_i = \log ( \frac{1}{1 + e^{-\beta} \cdot B } ) + \log ( \frac{e^{x_i}}{ \sum_{j=1}^{C}{e^{x_j}}} ), \\
\label{Equ:bg_cal_set1}
\end{equation}
\begin{equation}\small
B = \frac{e^{x_{0}}}{\sum_{j=1}^{C}{e^{x_j}}} = \frac{e^{x_{0}}}{\sum_{j=0}^{C}{e^{x_j}}} \cdot \frac{1}{1 - \frac{e^{x_{0}}}{\sum_{j=0}^{C}{e^{x_j}}}},
\label{Equ:bg_cal_set2}
\end{equation}
where $B$ is a monotonic function of the initial background score, $\frac{e^{x_{0}}}{\sum_{j=0}^{C}{e^{x_j}}}$.
In~\cref{Equ:bg_cal_set1},
when $\beta < 0$, its first item becomes more sensitive to the change of $B$ than the case of $\beta \geq 0$,
leading to a stronger \textit{attention} on background.

$\beta$ determines the contribution of background to detection results, and is dependent on the characteristics of datasets.
Generally, a dataset with more classes such as LVIS~\cite{lvis} contains fewer samples per class compared to a dataset with fewer classes such as MS COCO~\cite{coco}.
Detectors trained on datasets like LVIS~\cite{lvis} tend to have a more \textit{imprecise} discrimination of foreground class relative to the background.
Therefore, we are inclined to allocate \textit{more} attention to background scores,
making full use of the superior discriminability of classifier on the background,
thereby eventually rectify the ranking of foreground proposals, and delivering better detection results.
In this case, 
a negative $\beta$ is appropriate.
However, in practice, 
the statistical mean of the background is often too large to reflect the function of $\beta$ and hampers the calibration. Consequently, we set $\beta$ to the minimum value of the average logits that is always negative to provide an adaptive solution,
and the ablation study in~\cref{exp:ablation} validates its effectiveness.

\section{Experiments}

\subsection{Experimental Setup}

\textbf{Datasets and evaluation metric.}
We conduct experiments on the long-tail Large Vocabulary Instance Segmentation (LVIS)~\cite{lvis} dataset to validate the effectiveness of the proposed LogN.
We use the latest version v1.0, which provides precise bounding box and mask annotations for 1,203 categories.
Its \textit{training} set with around 100k images and 1.3M instances is employed for training and the \textit{val} set with around 19.8k images and 
244k instances is then taken for evaluation. 
Particularly, all categories are divided into three groups according to how many images they appear in training set:
\textit{rare} (1-10 images), \textit{common} (11-100 images), and \textit{frequent} ($>$100 images).
Hence for evaluation, apart from the commonly-used metric $\textbf{AP}^{b}$ (for object detection) and $\textbf{AP}$ (for instance segmentation),
the fine-grained metrics including $\textbf{AP}_r$ (for \textit{rare} classes), $\textbf{AP}_c$ (for \textit{common} classes), and $\textbf{AP}_f$ (for \textit{frequent} classes)
are also reported.

\textbf{Implementation details.}
Our LogN calibrates predictions during evaluation for pretrained baselines, which are implemented with MMDetection~\cite{mmdetection}
and identical to~\cite{seesawloss} in training details for a fair comparison. Concretely,
two-stage detectors Mask R-CNN~\cite{maskrcnn} and Cascade Mask R-CNN~\cite{cascade} with backbone ResNet-FPN~\cite{resnet,fpn} pretrained on ImageNet~\cite{imagenet} are employed and we train them using the 2x training schedule~\cite{mmdetection}, \ie the initial learning rate is 0.02 and decreases by 90\% after the 16th and 22nd epochs, respectively.
All models are optimized using SGD with a momentum of 0.9 and a weight decay of 0.0001 and a batch size of 16 on 8 GPUs
for 24 epochs.
Following the convention, scale jitter and horizontal flipping are used in training and no test time augmentation is used.
As in~\cite{seesawloss,equilibrium_loss}, we adopt normalized linear activation for both category and mask prediction.
In addition to the internal sampler that randomly samples training images,
the repetition factor sampler (RFS)~\cite{lvis} is also experimentally evaluated as the external sampler.
RFS oversamples categories appearing in less than 0.1\% of the total images and effectively enhances the overall AP.
Unless specific, all ablation studies and further analysis are conducted on Mask R-CNN with ResNet-101-FPN with RFS.

\begin{table*}[]
\small
\centering
\vspace{-3.5mm}
\caption{\textbf{Comparison with state-of-the-art methods on LVIS v1 dataset~\cite{lvis}.}
All methods are grouped according to their \textbf{framework} and training \textbf{sampler}, and we additionally report whether models perform \textbf{training} and the number of training \textbf{epochs}.
Mask: Mask R-CNN~\cite{maskrcnn};
Cascade: Cascade Mask R-CNN~\cite{cascade};
R50: ResNet-50-FPN~\cite{resnet,fpn};
R101: ResNet-101-FPN~\cite{resnet,fpn};
Random: no external sampler employed;
RFS: Repeat Factor Sampler~\cite{lvis};
MFS: Memory-augmented Feature Sampling~\cite{equilibrium_loss};
+: using decouple training pipeline~\cite{decouple};
-: merely post-hoc calibration.
}
\scalebox{0.87}{
\begin{tabular}{c|l|ccc|cc|ccc}
\toprule
\textbf{Framework} & \textbf{Method}              & \textbf{Sampler}& \textbf{Train}& \textbf{Epochs} & $\textbf{AP}^b$ & \textbf{AP} & $\textbf{AP}_r$ & $\textbf{AP}_c$ & $\textbf{AP}_f$ \\\midrule
\multirow{12}{*}{Mask-R50}
 & CE                                               & Random & \cmark & 24    & 19.9   & 19.0 &  1.3 & 16.7 & 29.3   \\
 & EQL~\cite{eqlv1}                                 & Random & \cmark & 24    & 22.5   & 21.6 &  3.8 & 21.7 & 29.2   \\
 & EQL v2~\cite{eqlv2}                              & Random & \cmark & 24    & 26.1   & 25.5 & 17.7 & 24.3 & 30.2   \\
 & Seesaw Loss~\cite{seesawloss}                    & Random & \cmark & 24    & 25.6   & 25.4 & 15.9 & 24.7 & 30.4    \\
 & LOCE~\cite{equilibrium_loss}                     & Random & \cmark & 24+6  & 24.0   & 23.8 &  8.3 & 23.7 & 30.7   \\
 & LogA~\cite{logit_adjustment,norcal}              & Random & \xmark & -     & 26.5   & 26.2 & 20.6 & 25.9 & 29.1   \\\cmidrule{2-10} 
 & \textbf{LogN}                                    & Random & \xmark & -     & \textbf{26.8}   & \textbf{26.5} & 19.2 & 25.7 & 30.4   \\
\cmidrule{2-10} 
 & CE                                             & RFS~\cite{lvis} & \cmark & 24    & 24.7 & 23.7 & 13.5 & 22.8 & 29.3  \\
 & Seesaw Loss~\cite{seesawloss}                  & RFS~\cite{lvis} & \cmark & 24    & 27.6 & 26.8 & 19.8 & 26.3 & 30.5    \\
 & LOCE~\cite{equilibrium_loss}                   & MFS~\cite{equilibrium_loss}             & \cmark & 24+6  & 27.4 & 26.6 & 18.5 & 26.2 & 30.7   \\
 & LogA~\cite{logit_adjustment,norcal}            & RFS~\cite{lvis} & \xmark & -     & 27.4 & 26.8 & 22.0 & 27.0 & 28.7   \\\cmidrule{2-10} 
 & \textbf{LogN}                                  & RFS~\cite{lvis} & \xmark & -     & \textbf{28.1} & \textbf{27.5} & 21.8 & 27.1 & 30.4 \\
\midrule
\multirow{14}{*}{Mask-R101} 
 & CE                                              & Random & \cmark &  24    & 21.7 & 20.6 & 0.8  & 19.3 & 30.7   \\
 & EQL~\cite{eqlv1}                                & Random & \cmark &  24    & 24.2 & 22.9 & 3.7  & 23.6 & 30.7   \\
 & BAGS~\cite{balanced_group_softmax}              & Random & \cmark &  24    & 26.4 & 25.6 & 17.3 & 25.0 & 30.1   \\
 & EQL v2~\cite{eqlv1}                             & Random & \cmark &  24    & 27.9 & 27.2 & 20.6 & 25.9 & 31.4   \\
 & Seesaw Loss~\cite{seesawloss}                   & Random & \cmark &  24    & 27.4 & 27.1 & 18.7 & 26.3 & 31.7   \\
 & LogA~\cite{logit_adjustment,norcal}             & Random & \xmark &  -     & 27.9 & 27.5 & 21.3 & 27.3 & 30.4   \\\cmidrule{2-10}
 & \textbf{LogN}                                   & Random & \xmark &  -     & \textbf{28.4} & \textbf{28.0} & 21.0 & 27.2 & 31.9     \\
\cmidrule{2-10} 
 & CE                                              & RFS~\cite{lvis} & \cmark &  24    & 26.6 & 25.5 & 16.6 & 24.5 & 30.6   \\
 & EQL~\cite{eqlv1}                                & RFS~\cite{lvis} & \cmark &  24    & 27.6 & 26.2 & 17.0 & 26.2 & 30.2   \\
 & BAGS~\cite{balanced_group_softmax}              & RFS~\cite{lvis} & \cmark &  24    & 26.5 & 25.8 & 16.5 & 25.7 & 30.1   \\
 & Seesaw Loss~\cite{seesawloss}                   & RFS~\cite{lvis} & \cmark &  24    & 28.9 & 28.1 & 20.0 & 28.0 & 31.8   \\
 & LOCE~\cite{equilibrium_loss}                    & MFS~\cite{equilibrium_loss}             & \cmark &  24+6  & 29.0 & 28.0 & 19.5 & 27.8 & 32.0   \\
 & LogA~\cite{logit_adjustment,norcal}             & RFS~\cite{lvis} & \xmark &  -     & 29.0 & 28.2 & 23.2 & 28.6 & 30.0   \\\cmidrule{2-10}
 & \textbf{LogN}                                   & RFS~\cite{lvis} & \xmark &  -     & \textbf{29.8} & \textbf{29.0} & 22.9 & 28.8 & 31.8   \\
\midrule
\multirow{13}{*}{Cascade-R101}
 & CE                                              & Random & \cmark &  24    & 25.5 & 22.6 & 2.4  & 22.0 & 32.2   \\
 & EQL~\cite{eqlv1}                                & Random & \cmark &  24    & 27.2 & 24.5 & 4.1  & 25.8 & 32.0   \\
 & BAGS~\cite{balanced_group_softmax}              & Random & \cmark &  24    & 31.5 & 27.9 & 19.6 & 27.7 & 31.6   \\
 & EQL v2~\cite{eqlv1}                             & Random & \cmark &  24    & 32.3 & 28.8 & 22.3 & 27.8 & 32.8   \\
 & Seesaw Loss~\cite{equilibrium_loss}             & Random & \cmark &  24    & 32.7 & 29.6 & 20.3 & 29.3 & 34.0   \\
 & LogA~\cite{logit_adjustment,norcal}             & Random & \xmark &  -     & 32.7 & 29.9 & 22.8 & 30.5 & 32.3   \\\cmidrule{2-10}
 & \textbf{LogN}                                   & Random & \xmark &  -     & \textbf{33.5} & \textbf{30.6} & 22.7 & 30.7 & 34.0   \\
\cmidrule{2-10} 
 & CE                                              & RFS~\cite{lvis} & \cmark &  24    & 30.3 & 27.0 & 16.6 & 26.7 & 32.0   \\
 & EQL~\cite{eqlv1}                                & RFS~\cite{lvis} & \cmark &  24    & 30.4 & 27.1 & 17.0 & 27.2 & 31.4   \\
 & BAGS~\cite{balanced_group_softmax}              & RFS~\cite{lvis} & \cmark &  24    & 30.2 & 27.0 & 16.9 & 26.9 & 31.7   \\
 & Seesaw Loss~\cite{equilibrium_loss}             & RFS~\cite{lvis} & \cmark &  24    & 32.8 & 30.1 & 21.4 & 30.0 & 33.9   \\
 & LogA~\cite{logit_adjustment,norcal}             & RFS~\cite{lvis} & \xmark &  -     & 32.7 & 30.0 & 23.9 & 30.8 & 31.7   \\\cmidrule{2-10}
 & \textbf{LogN}                                   & RFS~\cite{lvis} & \xmark &  -     & \textbf{33.8} & \textbf{30.9} & 23.9 & 31.1 & 33.8    \\
\bottomrule
\end{tabular}
}
\label{tab:ltod_framework}
\end{table*}

\subsection{Comparison with State of the Arts}\label{sota}
To display the effectiveness of LogN on long-tail object detection, we compare LogN with the state-of-the-art methods on LVIS v1.0 dataset.
We present the results with and without external sampler to show the effectiveness of LogN at multiple circumstances.
As shown in~\cref{tab:ltod_framework}, our LogN outperforms all state-of-the-art methods with and without external sampler on various detection frameworks.
Specifically, with both Cascade Mask R-CNN and RFS, LogN outperforms Seesaw~Loss~\cite{seesawloss} by 1.0\% and 0.8\% on $\textbf{AP}^b$ and \textbf{AP}, respectively. 
LogN also achieves higher $\textbf{AP}_r$ and $\textbf{AP}_c$ than Seesaw~Loss by 2.5\% and 1.1\% and is only 0.1\% poorer than it on $\textbf{AP}_f$, 
which demonstrates our method greatly reduces the bias in long-tail learning.
Notably, similar to LogA~\cite{logit_adjustment,norcal}, LogN is executed solely in the testing phase, eliminating the requirement for a time-consuming and laborious training process as in~\cite{eqlv1,eqlv2,equilibrium_loss,seesawloss}.
More importantly, our method further consistently outperforms LogA, especially with the external sampler.
In particular, with RFS, LogN outperforms LogA by 1.1\% $\textbf{AP}^b$ on Cascade-R101, 0.8\% $\textbf{AP}^b$ on Mask-R101 and 0.7\% $\textbf{AP}^b$ on Mask-R50, respectively.
This indicates that our LogN is more compatible with external samplers and is indeed more adaptive to the alternation of label distribution.

\subsection{Ablation Study} \label{exp:ablation}

\begin{wraptable}[11]{r}{.5\textwidth}
    \vspace{-8mm}
    \tabcaption{\textbf{Study on components of LogN.}
    $\mu $: mean. $\sigma $: standard deviation. $\beta $: background calibration scalar.}
    \vspace{-3mm}
    \centering
    \begin{tabular}{ccc|cc|ccc}
    \toprule
    $\mu $  &  $\sigma $  & $\beta$  & $\textbf{AP}^b$ & \textbf{AP} & $\textbf{AP}_r$ & $\textbf{AP}_c$ & $\textbf{AP}_f$ \\\midrule
            &        &         &    27.5          &  26.8         & 17.4 & 26.2 & 31.7  \\
     \cmark &        &         &    28.7         &  28.2         & 21.5 & 27.9 & 31.5   \\ 
            & \cmark &         &    29.1          &   28.5        & 21.4 & 28.3 & 31.9   \\ 
            &        &  \cmark &    29.0          & 28.0          & 19.3 & 27.4 & 32.4   \\ 
     \cmark & \cmark &  \cmark &    \textbf{29.8} & \textbf{29.0} & 22.9 & 28.8 & 31.8   \\
    \bottomrule
    \end{tabular}
    \label{tab:ablation1}
\end{wraptable}

\textbf{Study on components.} We conduct ablation studies on the components in our LogN, \eg, mean $\mu $, standard deviation $\sigma $ and background calibration scalar $\beta$.
As shown in~\cref{tab:ablation1}, $\mu $ pushes the performance on $\textbf{AP}_r$ from 17.4\% to 21.5\%. This performance gain also contributes to significant increases on both $\textbf{AP}^b$ and $\textbf{AP}$. 
In comparison, $\sigma $ delivers greater improvements on both common and frequent classes, and also leads to better final results. This indicates that both statistics bring promising effects, yet surprisingly, compared to logit mean, which better aligns with the training label distribution, dividing variance corrects the long-tail bias more effectively and thoroughly.
We further validate the effectiveness of $\beta $, \ie the fourth line.
Its impressive results, especially on $\textbf{AP}^b$ and $\textbf{AP}_r$, suggesting the background calibration greatly alleviates the impact caused by foreground-background class imbalance.
Finally, LogN achieves the best results with all three components included.

\textbf{Study on selection of $\beta$.} We carry out ablation studies on the selection of $\beta$ in LogN. Based on our analysis, we find it is suitable to assign it with a negative value. In~\cref{tab:ablation_beta}, we investigate how its value influence the performance quantitatively. We report the results with the original statistical mean value of the background category, $\mu_{0} $~($\approx13.44$), and unsurprisingly, we find its performance is the worst.
When using $\mu^+_{\max} $~($\approx3.11$), it still performs bad under each metric. However, when employing $\mu^+_{\mathrm{avg}} $~($\approx-1.09$) and $\mu^+_{\min} $~($\approx-5.35$), the performance comes to a higher level.
Additionally, we try to take $\beta$ as a constant value, and manually select two values (\eg -4 and -12).
Generally, the comparison in~\cref{tab:ablation_beta} suggests that $\beta$ has a certain value range and,
$\mu^+_{\min} $ is suitable for it as an \textit{adaptive} solution.

\textbf{Study on object detection frameworks.} We conduct experiments on various detectors with diverse backbones. As shown in~\cref{tab:ltod_framework}, LogN achieves the state-of-the-art performance under multiple metrics. Furthermore, we observe incremental gains by scaling up models.

\begin{table*}[]
    \parbox{0.42\textwidth}{
        \small
        \centering
        \setlength{\tabcolsep}{0.6mm}
        \caption{\textbf{Study on the selection of $\beta$.}
        $\mu_{0} $: mean of the background logit. $\mu^+_{\max} $: maximum of the foreground logits. $\mu^+_{\mathrm{avg}} $: average of the foreground logits. $\mu^+_{\min} $: minimum of the foreground logits.}
        \begin{tabular}{c|cc|ccc}
        \toprule
         $\beta$  & $\textbf{AP}^b$ & \textbf{AP} & $\textbf{AP}_r$ & $\textbf{AP}_c$ & $\textbf{AP}_f$ \\\midrule
         $-4$                    &   29.6         &  28.8         &  22.9  & 28.8 &  31.3    \\
         $-12$                   &   29.3         &  28.3         &  20.8  & 28.0 &  31.9    \\
         $\mu_{0} $              & 12.1           &  12.5         &  14.1  & 14.5 &  9.6  \\
         $\mu^+_{\max} $         &   25.5         &  25.4         &  20.9  & 25.7 &  27.0   \\
         $\mu^+_{\mathrm{avg}} $ &  29.4          &  28.7         &  23.5  & 29.0 &  30.7   \\
         $\mu^+_{\min} $         &  \textbf{29.8} & \textbf{29.0} &  22.9  & 28.8 &  31.8   \\
        \bottomrule
        \end{tabular}
        \label{tab:ablation_beta}
        \vspace{-2mm}
    }
    \qquad
    \parbox{0.5\textwidth}{
        \small
        \centering
        \setlength{\tabcolsep}{0.2mm}
        \caption{Comparison with different methods on ImageNet-LT dataset~\cite{imagenetlt}.}
        \begin{tabular}{l|ccc|c}
        \toprule
        \textbf{Method} & \textit{Many} & \textit{Medium} & \textit{Few} & \textbf{Overall} \\
        \midrule
        Softmax                      & 65.9 & 37.5 &  7.7 & 44.4\\
        Focal Loss~\cite{focal}      & 64.5 & 36.3 &  7.8 & 43.3 \\
        cRT~\cite{decouple}          & 61.8 & 46.2 & 27.4 & 49.6 \\
        $\tau$-norm~\cite{decouple}  & 59.1 & 46.9 & 30.7 & 49.4 \\
        LWS~\cite{decouple}          & 60.2 & 47.2 & 30.3 & 49.9 \\
        EQL~\cite{eqlv1}             & 61.7 & 42.5 & 13.8 & 46.0 \\
        B-Softmax~\cite{balanced_meta_softmax} & 62.2 & 48.8 & 29.8 & 51.4 \\
        Seesaw Loss~\cite{seesawloss} & 67.1 & 45.2 & 21.4 & 50.4 \\
        LogA~\cite{logit_adjustment} & 63.4 & 47.9 & 31.5 & \textbf{51.6} \\
        \midrule
        \textbf{LogN} & 59.1 & 50.3 & 35.0 & \textbf{51.6} \\
        \bottomrule
        \end{tabular}
        \label{tab:imagenet_lt}
        \vspace{-2mm}
    }
\end{table*}

\subsection{Further Analysis} \label{exp:analysis}

\begin{wrapfigure}[9]{r}{.5\textwidth}
    \centering
    \vspace{-6mm}
    \includegraphics[width=1.\linewidth]{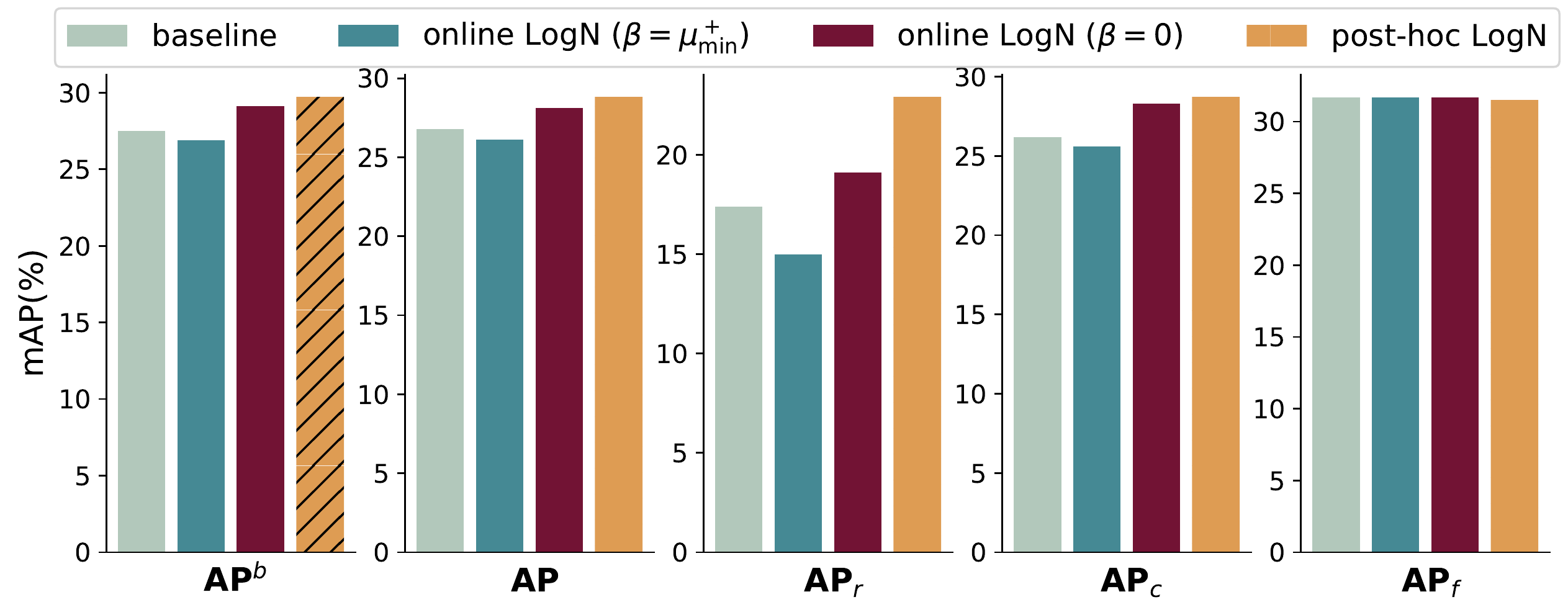}
    \vspace{-8mm}
    \caption{Per-group results of online and post-hoc LogN and their baseline.}
    \label{online}
    \vspace{-5mm}
\end{wrapfigure}

\textbf{Why not calibrate online?}
As analyzed in~\cref{post-hoc}, there is also an online equivalent form of our LogN.
In particular, for each batch of training logits, we first accumulate the global statistics as in BN~\cite{bn}, 
and then apply them to \textit{scale and shift} the logits as in~\cref{Equ:invnorm}, 
before calculating the objective, \eg cross-entropy loss.
Without any processing of the logits during inference, the per-group detection results 
of two instantiations ($\beta=\mu^+_{\min}$ and $\beta=0$) of the online LogN are shown in~\cref{online}.
Interestingly, the former degrades the performance of baseline, 
suggesting a significant difference of calibration on background between training and testing phases;
whereas the latter enhances baseline's performance, validating our analysis.
However, there is still a small performance gap between it and our post-hoc LogN,
because of the \textit{imprecision} of statistics in the \textit{early} training period,
which is why we employ the post-hoc LogN as the final solution.

\begin{figure}[t!]
    \centering
    \subfloat[Baseline]{
        {\includegraphics[width=0.31\columnwidth]{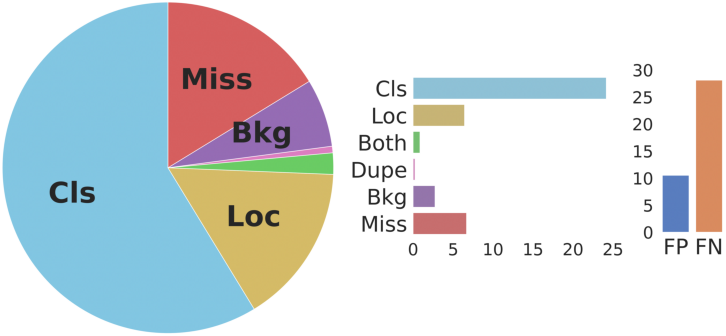}
        \label{Fig:Baseline_tide}}}
    \qquad
        \subfloat[Our LogN]{
        {\includegraphics[width=0.31\columnwidth]{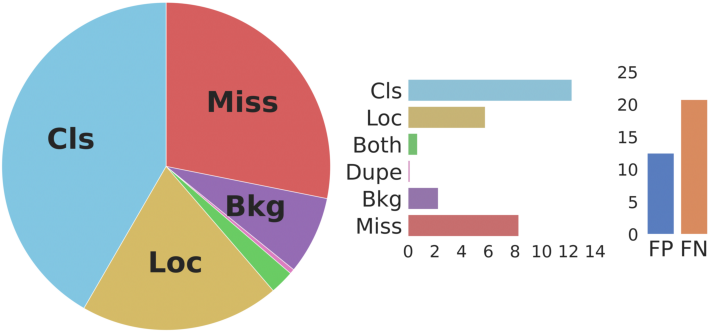} 
        \label{Fig:LogN_tide}}}
\vspace{-2mm}
\caption{
    \textbf{Error analysis for LogN and its baseline.}
    According to TIDE~\cite{tide}, the figure shows the six types of errors, 
    the pie chart shows the relative contribution of each error, 
    while the bar plots show their absolute contribution.
    Cls: classification error; 
    Loc: localization error; 
    Both: both cls and loc error;
    Dupe: duplicate detection error;
    Bkg: background error;
    Miss: missed ground truth error. 
    FP: false positive;
    FN: false negative.
}
\label{Fig:error}
\vspace{-5mm}
\end{figure}

\textbf{How LogN calibrate detections?}
To investigate the influence of LogN on the detection results at a finer scale, we conduct an error analysis of the detections before and after the execution of LogN with the recently proposed TIDE~\cite{tide}.
As depicted in~\cref{Fig:error}, LogN greatly reduces the \textit{classification error} in long-tail object detection, with a large decrease on \textit{false negative} rate.
However, among these improvements, LogN leads to slightly higher false positive rate. We speculate it is because the negative samples also have low confidence as well as the candidates of rare classes, however, the operation of subtracting statistical mean in LogN decreases the confidence in each candidate, 
mixing up the candidates originally from various score bands, 
and causing an increase on false positive rate.

\subsection{Long-tail Image Classification}

We further conduct experiments on ImageNet-LT~\cite{imagenetlt} to demonstrate the competitiveness and versatility of LogN.
ImageNet-LT is a long-tail version of the large-scale object classification dataset ImageNet~\cite{imagenet},
and contains 115.8k images of 1,000 categories, with class cardinality ranging from 5 to 1,280.
Following~\cite{imagenetlt}, we also divide the categories of ImageNet-LT into three groups according to the number of samples in each class:
\textit{Few} (1-19 images), \textit{Medium} (20-100 images), and \textit{Many} ($>$100 images).
Thus for evaluation, we report top-1 accuracy of the \textbf{Overall} evaluation results, as well as results on each group.
The experimental implementation and mainly reported results follow~\cite{seesawloss}.

It can be observed in~\cref{tab:imagenet_lt} that LogN yields 51.6\% top-1 overall accuracy, which is competitive among diverse previous methods.
Notably, the performance of LogN is roughly on par with some recent prior-based methods~\cite{balanced_meta_softmax,logit_adjustment} that have been proven to be theoretically optimal. 
This fully demonstrates that in a simple image classification task with only random sampling, LogN is still able to precisely \textit{capture} and effectively \textit{de-bias} the long-tail training label distribution that is consistent with the prior label distribution.
Moreover, our method also outperforms EQL~\cite{eqlv1} and Seesaw Loss~\cite{seesawloss} devised for object detection and instance segmentation. This is in line with~\cref{sota} and well showcases the superior generalization of LogN.
It is noteworthy that, contrary to the results in~\cref{sota}, LogN offers better performance on tail classes and worse performance on head classes compared to LogA~\cite{logit_adjustment}.
We speculate that this stems from the behavioral difference of variance, which is indeed introduced by the differences in datasets and tasks.
For classification in ImageNet~\cite{imagenet}, variance actually balances the magnitudes of logits to some degree, which equals to give classifiers more balanced decision boundaries just like $\tau$-norm~\cite{decouple}, as verified by their similar many-shot accuracy.
For more details, please refer to supplementary material.

\section{Conclusion}

In this paper, we propose Logit Normalization (LogN), a simple technique to self-calibrate the classified logits of object detectors like batch normalization.
Extensive experiments on long-tail object detection and classification demonstrate its impressive effectiveness and versatility.
Overall, LogN is based on statistics and thus sufficient to dynamically adapt to various distributions and networks. 
It is also independent of training and tuning and can be directly applied without any bells and whistles.

\appendix
\section*{Appendix}

\counterwithin{figure}{section}
\counterwithin{table}{section}
\section{More Implementation Details}

On Fig. 1 and Fig. 2, all data are captured from the \textit{training} set of LVIS v1~\cite{lvis} via Mask R-CNN~\cite{maskrcnn} with ResNet-101-FPN~\cite{resnet,fpn}. And all results on Fig. 3 and Fig. 4 are obtained from the same detector evaluated on the \textit{val} set.

For experiments on long-tail object detection, we adopt multi-scale with horizontally flip augmentation during training, following the practices in mmdetection~\cite{mmdetection} and prior efforts~\cite{seesawloss,eqlv1,eqlv2,seesawloss}. In specific, during training, we randomly resize the shorter edge of the image within {640, 672, 704, 736, 768, 800} pixels and maintain the longer edge smaller than 1333 pixels without changing the aspect ratio. During testing, we then employ single-scale testing with image size of 1333 × 800 pixels.

For experiments on long-tail classification, all implementation details follow~\cite{seesawloss}.
Specifically, we employ ResNeXt-50~\cite{resnext} as the backbone network and SGD with momentum of 0.9 and weight decay of $5 \times 10^{-4}$ as the optimizer to end-to-end train the model for 90 epochs.
In line with Seesaw Loss~\cite{seesawloss}, we also adopt the normalized linear activation for Logit Adjustment (LogA)~\cite{logit_adjustment}  and our LogN.

\section{More Analysis}

\subsection{How $\mu, \sigma$ and $\beta$ function?}
\begin{figure}[t!]
    \centering
    \subfloat[Baseline]{
        {\includegraphics[width=0.3\columnwidth]{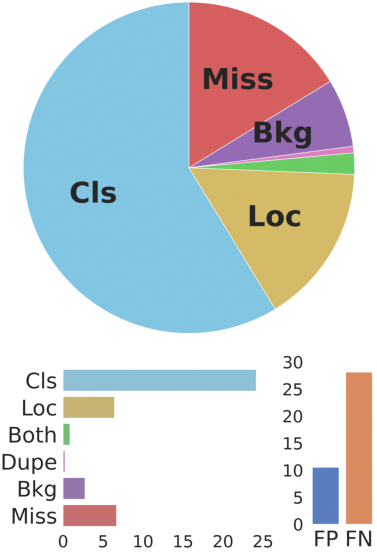}
        \label{Fig:Baseline_tide_sup}}}
    \subfloat[Baseline+$\mu$]{
        {\includegraphics[width=0.3\columnwidth]{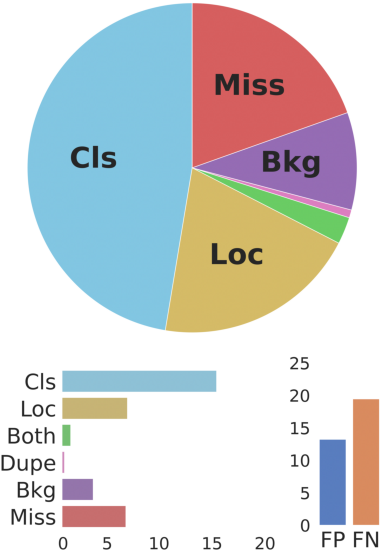} 
        \label{Fig:mu_tide}}}
    \subfloat[Baseline+$\sigma$]{
        {\includegraphics[width=0.3\columnwidth]{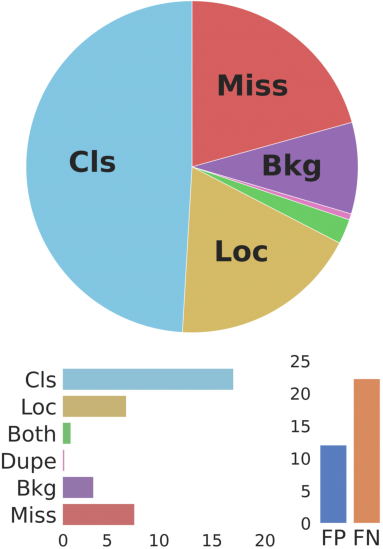} 
        \label{Fig:var_tide}}}
    \\
    \subfloat[Baseline+$\beta$]{
        {\includegraphics[width=0.3\columnwidth]{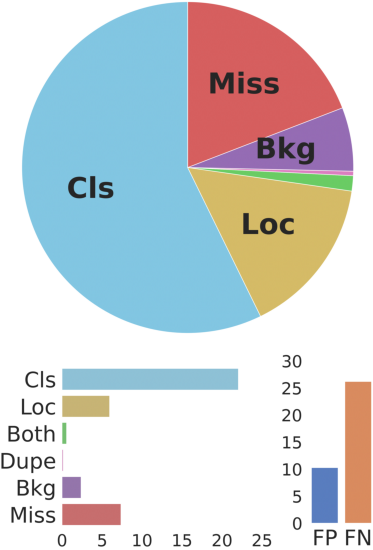} 
        \label{Fig:beta_tide}}}
    \qquad
    \subfloat[Baseline+$\mu, \sigma, \beta$]{
        {\includegraphics[width=0.3\columnwidth]{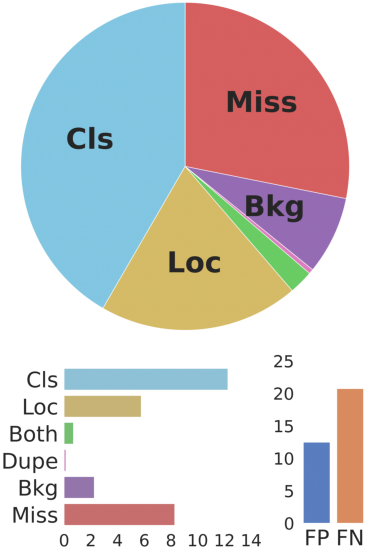} 
        \label{Fig:LogN_tide_sup}}}
\vspace{-2mm}
\caption{
    \textbf{Error analysis for components of LogN.}
    According to TIDE~\cite{tide}, the figure shows the six types of errors, 
    the pie chart shows the relative contribution of each error, 
    while the bar plots show their absolute contribution.
    Cls: classification error; 
    Loc: localization error; 
    Both: both cls and loc error;
    Dupe: duplicate detection error;
    Bkg: background error;
    Miss: missed ground truth error. 
    FP: false positive;
    FN: false negative.
}
\label{Fig:error_4}
\vspace{-5mm}
\end{figure}

To further clarify the functioning mechanism of each component within LogN, we also conduct error analysis of detections with and without each component, just like our ablation study. The results are shown in~\cref{Fig:error_4}.
Comparing the results from~\cref{Fig:Baseline_tide_sup} and~\cref{Fig:mu_tide}, the logit mean leads to a great decrease on classification error with a slight increase on the false positive, illustrating it mostly improves the accuracy of model prediction. 
Comparing the results from~\cref{Fig:Baseline_tide_sup} and~\cref{Fig:var_tide}, the statistical variance also reduces the classification error. In quantity, it reduces it from 24.17 dAP to 17.10 dAP.
As for the last comparison, the presence of $\beta$ \textit{consistently reduces all kinds of errors}, revealing the significance of balancing the foreground-background imbalance. Quantitatively, it reduces the false positive rate from 10.52 dAP to 10.30 dAP, the false negative rate from 28.10 dAP to 26.21 dAP, and the classification error from 24.17 dAP to 22.05 dAP.

\subsection{How $\beta$ influences background?}
In this section, we illustrate how $\beta $ amplify the effect of the initial background score.

Assuming $y(s_{b}, s_{f}, \beta) = \log ( \frac{1}{1 + e^{-\beta} \cdot B( s_b) } ) + \log (s_f) $, where $
B( s_b) = \frac{s_b}{1 - s_b}$ and $s_b, s_f \in (0,1)$. Then, we calculate the changing rate of $y$, that is, $k(\Delta s_b, s_b, s_f, \beta)$, on the component $s_b$:
\begin{equation}
\begin{aligned}
 & k(\Delta s_b, s_b, s_f, \beta) = \frac{y(s_b + \Delta s_b, s_f, \beta) - y(s_b , s_f, \beta)}{\Delta s_b} \\
 &= \frac{-1}{\Delta s_b} \log(1 + e^{-\beta} \frac{B(s_b + \Delta s_b) - B(s_b)}{1+e^{-\beta} B(s_b)}) \\
 &= \frac{-1}{\Delta s_b} \log(1 + e^{-\beta} \frac{ \frac{\Delta s_b}{(1-s_b)(1-s_b-\Delta s_b)}}{1+e^{-\beta} B(s_b)}). \\
\end{aligned}
\label{Equ:sb_sf}
\end{equation}
Therefore, according to~\cref{Equ:sb_sf}, when $\beta < 0$, the inequality $|k(\Delta s_b, s_b, s_f, \beta)| > |k(\Delta s_b, s_b, s_f, 0)|$ holds. Meanwhile, $\beta$ does not influence the change on $s_f$, and $y(s_{b}, s_{f}, \beta)$ is continuous and differentiable.

As for the task of classification in object detection, $s_b$ from~\cref{Equ:sb_sf} is the probability of background prediction, that is, $\Pr(c=0|x; \theta)$. $s_f$ indicates the conditional probability of one class over foreground categories, that is, $\Pr(c=i|c>0, x; \theta)$. Thus, in the confidence ranking, comparing two candidates predicting the same category, the background confidence are crucial when $\beta < 0$. In a nutshell, $\beta < 0$ amplifies the effect of the initial background score on the classification of object detection.

\subsection{Why calculate statistics on all samples like BN?}\label{whyallsample}
\begin{figure}
\centering
\includegraphics[width=0.8\columnwidth]{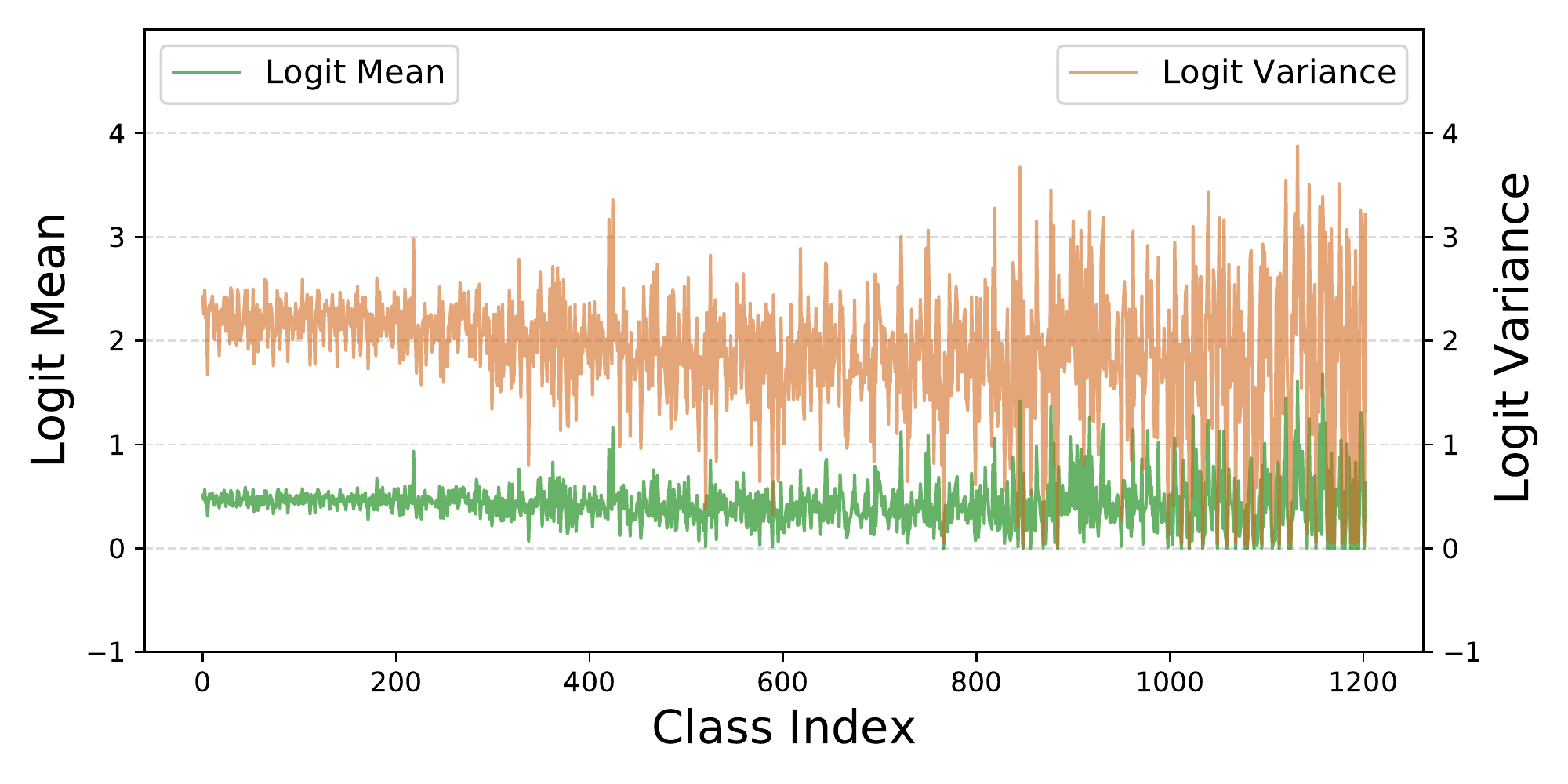}
\vspace{-2mm}
\caption{Statistics accumulated from only positive samples.}
\label{Fig:pos_sta2}
\vspace{-5mm}
\end{figure}

As emphasized in main body, the statistics for each class are calculated over \textit{all} samples, which is consistent with that in BN~\cite{bn}. 
However, 
another intuitive way to accumulate statistics for each category is to use only \textit{positive} samples.
Specifically, for one proposal, only its logit on the correctly predicted class is accumulated while the logits on the other classes are masked.
Problematically, this operation \textit{neglects the margins} between classes and leads to a more \textit{unstable} distribution, shown in~\cref{Fig:pos_sta2}.
These statistics cannot represent the original long-tail distribution, and their trends are more uniform. Additionally, they have much more fluctuations on the tail classes, causing unstable estimations of statistics on these categories.
In general, with such statistics for normalization, the logits on head classes cannot be correctly inhibited, leading to a worse performance for calibration.

In contrast, our strategy, \ie using all logits of every proposal for calculating the statistics,
properly \textit{preserves the margins} between classes. Therefore, statistics on tail classes benefited from the abundant logits of proposals become \textit{stable} and \textit{robust}.

\subsection{Why results in detection and classification are inconsistently improved?}
Throughout our experimental results, we find that compared to LogA~\cite{logit_adjustment},
LogN is worse on tail classes and better on head classes in object detection, yet is better on tail classes and worse on head classes in classification.

We attribute this to the behavioral differences in variance across tasks and datasets.
As shown in~\cref{tab:ablation_imagenet_lt}, for image classification on ImageNet-LT~\cite{imagenetlt}, 
it is evident that, both statistics deliver significant performance gains, \ie +5.6\% and +6.5\% relatively to baseline (training with Softmax).
But among the gains of performance, 
$\mu$ enhances the performance of all classes simultaneously, 
while $\sigma$ is more of a trade-off, reducing the accuracy on many-shot classes
yet improving the accuracy on both medium- and few-shot classes.
Thus, we speculate that the variance acts like \textit{re-balancing} the performance on each group of classes, just like $\tau$-norm~\cite{decouple}, which specializes in \textit{providing more balanced classifier weight magnitudes}.

Object detection on LVIS~\cite{lvis} is a much more challenging task compared to classification. 
The complex coupling of multiple modules and samplers as well as the task of distinguishing the background proposal make it difficult to use statistics to dynamically estimate the training label distribution.
The problem is that, although the two statistics provide efficient and comparable boosts, as shown in Tab. 2, the variance \textit{fails} in effectively equalizing the performance on the different groups of classes.
Specifically, the magnitude of variance is too \textit{small} to inhibit the confidence on head classes and increase that on tail classes.
Therefore, we try to square the standard deviation of LogN.
On Mask R-CNN 101 with RFS~\cite{lvis}, we find that after this operation, the performance of LogN on rare classes can achieve 23.1\% which is very close to that of LogA, \ie 23.2\%. However, it also leads to a performance drop on the total mAP, \ie -0.5\%.
Thus, we do not include such additional operations in our method.

\begin{table}[]
    \small
    \centering
    \caption{Study on components of LogN on ImageNet-LT~\cite{imagenetlt}.}
    \begin{tabular}{cc|ccc|c}
    \toprule
    $\mu$ & $\sigma$ & \textit{Many} & \textit{Medium} & \textit{Few} & \textbf{Overall} \\
    \midrule
    \cmark &   & 65.8 & 45.0 & 22.8 & 50.0 \\
     & \cmark & 63.0 & 48.4 & 25.9 & 50.9 \\
    \cmark & \cmark & 59.1 & 50.3 & 35.0 & \textbf{51.6} \\
    \bottomrule
    \end{tabular}
    \label{tab:ablation_imagenet_lt}
    \vspace{-2mm}
\end{table}

\clearpage
\bibliographystyle{splncs04}
\bibliography{main}
\end{document}